\documentclass{article}

\usepackage[preprint]{neurips_2025}

\usepackage[utf8]{inputenc} 
\usepackage[T1]{fontenc}    
\usepackage{microtype}      

\usepackage{amsmath}        
\usepackage{amsfonts}       
\usepackage{nicefrac}       

\usepackage{booktabs}       
\usepackage{multirow} 
\usepackage{array}
\usepackage{tabularx}
\usepackage[table,xcdraw]{xcolor} 

\usepackage{graphicx}
\usepackage{subcaption}
\usepackage{fancyhdr}

\usepackage{hyperref}       
\usepackage{url}            

\title{Lost in Translation and Noise: A Deep Dive into the Failure Modes of VLMs on Real-World Tables}

\author{
  Anshul Singh \\
  Indian Institute of Science, Bangalore \\
  \texttt{anshulsinghchambial@gmail.com} \\
  \And
  Rohan Chaudhary \\
  Panjab University, Chandigarh \\
  \texttt{chaudhary18.rohan@gmail.com}\\
  \AND
  Gagneet Singh\\
  Panjab University, Chandigarh \\
  \texttt{gagneet5647@gmail.com} \\
  \And
  Abhay Kumar \\
  Panjab University, Chandigarh \\
  \texttt{abhaykumar.connect@gmail.com} \\
}

\begin{document}
\maketitle

\begin{abstract}
The impressive performance of VLMs is largely measured on benchmarks that fail to capture the complexities of real-world scenarios. Existing datasets for tabular QA, such as WikiTableQuestions and FinQA, are overwhelmingly monolingual (English) and present tables in a digitally perfect, clean format. This creates a significant gap between research and practice. To address this, we present \textbf{MirageTVQA}, a new benchmark designed to evaluate VLMs on these exact dimensions. Featuring nearly 60,000 QA pairs across 24 languages, MirageTVQA challenges models with tables that are not only multilingual but also visually imperfect, incorporating realistic noise to mimic scanned documents. Our evaluation of the leading VLMs reveals two primary failure points: a severe degradation in performance (over 35\% drop for the best models) when faced with visual noise and a consistent English-first bias where reasoning abilities fail to transfer to other languages. MirageTVQA provides a benchmark for measuring and driving progress towards more robust VLM models for table reasoning. The dataset and the code are available at: \url{https://github.com/anshulsc/MirageTVQA}.
\end{abstract}

\section{Introduction}
\label{sec:introduction}

Tables are the backbone of information storage in countless domains, from financial reports and scientific papers to enterprise databases and healthcare records. The ability to comprehend and reason over these semistructured data is an important skill for LLMs to act as an AI agent. Traditionally, answering questions from table images involved a cascading pipeline: first, an Optical Character Recognition (OCR) engine extracts text, which is then fed to a language model for reasoning. However, this multi-step process is highly susceptible to error propagation, where a single OCR mistake can invalidate the entire result. Recent work has shown that end-to-end vision language models (VLMs) can outperform these brittle OCR+LLM pipelines by jointly processing visual and textual information \citep{multimodaltable24}.

This positions VLMs as a promising direction for robust table understanding. However, their true potential is hampered by a critical blind spot in existing evaluation benchmarks. Research has progressed along two largely separate tracks. On one hand, most benchmarks that explore complex reasoning, like FinQA \citep{chen2021finqa}, remain fundamentally \textbf{text-based}, ignoring the visual aspect of tables as they appear in documents. On the other hand, the few benchmarks that incorporate visual elements are almost exclusively \textbf{English-centric}. Crucially, to our knowledge, no existing benchmark addresses these two critical real-world challenges, visual complexity and linguistic diversity, \textbf{simultaneously}.

To fill this gap, we introduce MirageTVQA. Our primary contributions are twofold.
\begin{itemize}
    \item We construct MirageTVQA, the first large-scale visual question-answering benchmark designed to test VLM reasoning on nearly 60,000 QA pairs. It integrates massive multilingual support (24 languages) with visually realistic table images.
    \item We conduct an extensive empirical evaluation of leading open-source VLMs, providing a comprehensive analysis of their performance across languages and their robustness against visual degradation.
\end{itemize}

\section{Related Work} 
\label{sec:related_work}

Our work builds upon and addresses the limitations of several lines of research in table understanding.

\textbf{Text-Based Table Question Answering.} A significant body of work has focused on reasoning over the textual content of tables. Benchmarks such as Spider \citep{yu2018spider}, WikiTableQuestions \citep{pasupat2015compositional}, TAT-QA \citep{ZhuLHWZLFC20}, TableBench \citep{tablebench24}, and MIMO-Table \citep{MiMoTable} frame the task as text-to-SQL or text-to-answer. These approaches typically feed models linearized HTML or Markdown renditions of tables, completely sidestepping the visual modality. While important for evaluating textual reasoning, they do not test a model's ability to process tables as they are found in documents.

\textbf{Multi-modal Table Question Answering.} More recent efforts have extended this research into the multi-modal domain. Work such as MMTab \citep{multimodaltable24} and MTabVQA \citep{singh2025mtabvqa} frame the task as visual table question answering, requiring models to reason over table images. However, these benchmarks have two key limitations: they remain monolingual (English-only), and the visual table images are synthetically generated and clean, lacking the real-world noise and artifacts often present in scanned documents or photographs.

\textbf{Multilingual Table Question Answering.} A third line of research has attempted to address the multilingual gap. Benchmarks like M3TQA \citep{shu2025m3tqa} provide datasets for evaluation across multiple languages. Nevertheless, these efforts are often limited to a small number of language pairs (e.g., Chinese–English) and, critically, focus solely on text-based table representations, thereby inheriting the limitations of the first category.
Our work is first to fill this gap by combining 20+ languages across diverse domains (scientific, financial, and general knowledge), evaluating a broader set of 10 reasoning types, introducing realistic visual noise to table images to create a benchmark to test VLM's tabular understanding.

\section{MirageTVQA Benchmark}

\subsection{Data Collection and Table Translation}
\textbf{Source Table Collection and Filtering.} We begin by collecting a diverse set of 3000 English tables from four primary sources: Wikipedia (WikiSQL \citep{wikisql17}), financial documents (FinQA \citep{chen2021finqa}), scientific papers from arXiv, and GitHub. To ensure the suitability of tables for translation, we filtered the tables based on the median word character count per cell and selected a representative subset for translation. This process filtered our initial set down to 250 seed tables for the multilingual generation phase: 50 from arXiv, 100 from Wikipedia, and 100 from other sources, respectively. This curated set ensures a balanced and high-quality foundation for our benchmark.

With this curated seed, we create our multilingual Table corpus using a \textbf{translate-refine-filter} pipeline. For each of the 30 selected target languages, we: (1) perform an initial translation of all textual content using \textbf{Qwen3-32B} (\cite{yang2025qwen3}); (2) use a powerful LLM (Gemini 2.5 Pro \citep{gemini_2024}) to refine the translation by cross-referencing the original English table for context and data integrity; (3) back-translate the refined table to English; and (4) filter out the languages that had low-quality translation on the basis of back-translation BLEU score (\cite{bleu}), resulting in translated tables in 24 diverse languages.

\begin{figure*}[b!]
  \centering
  \scriptsize  
  \begin{subfigure}[b]{0.35\textwidth}
    \centering
    \includegraphics[width=\textwidth]{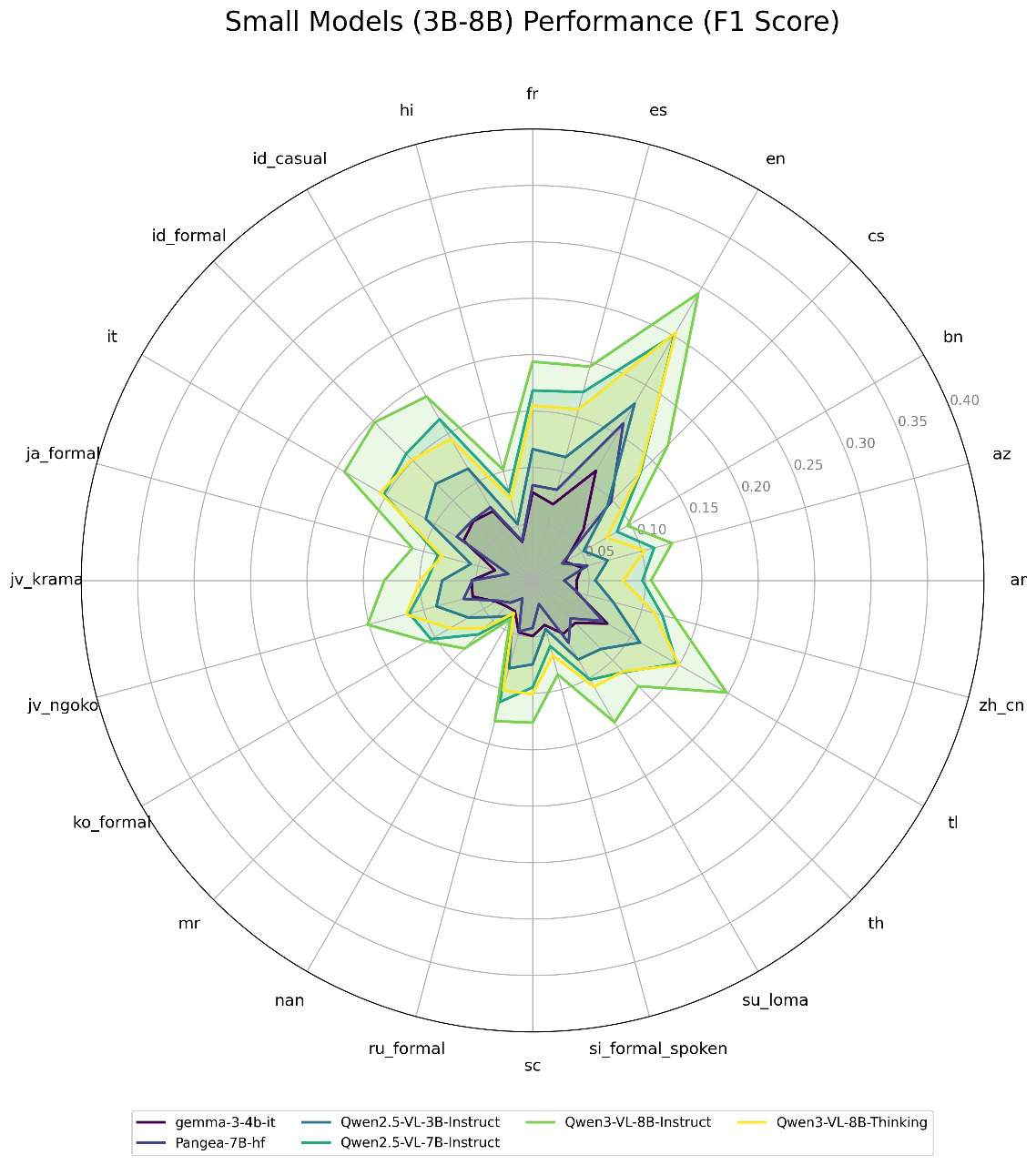}
    \caption{Small (3B-8B)}
    \label{fig:radar_small}
  \end{subfigure}
  \hfill
  \begin{subfigure}[b]{0.35\textwidth}
    \centering
    \includegraphics[width=\textwidth]{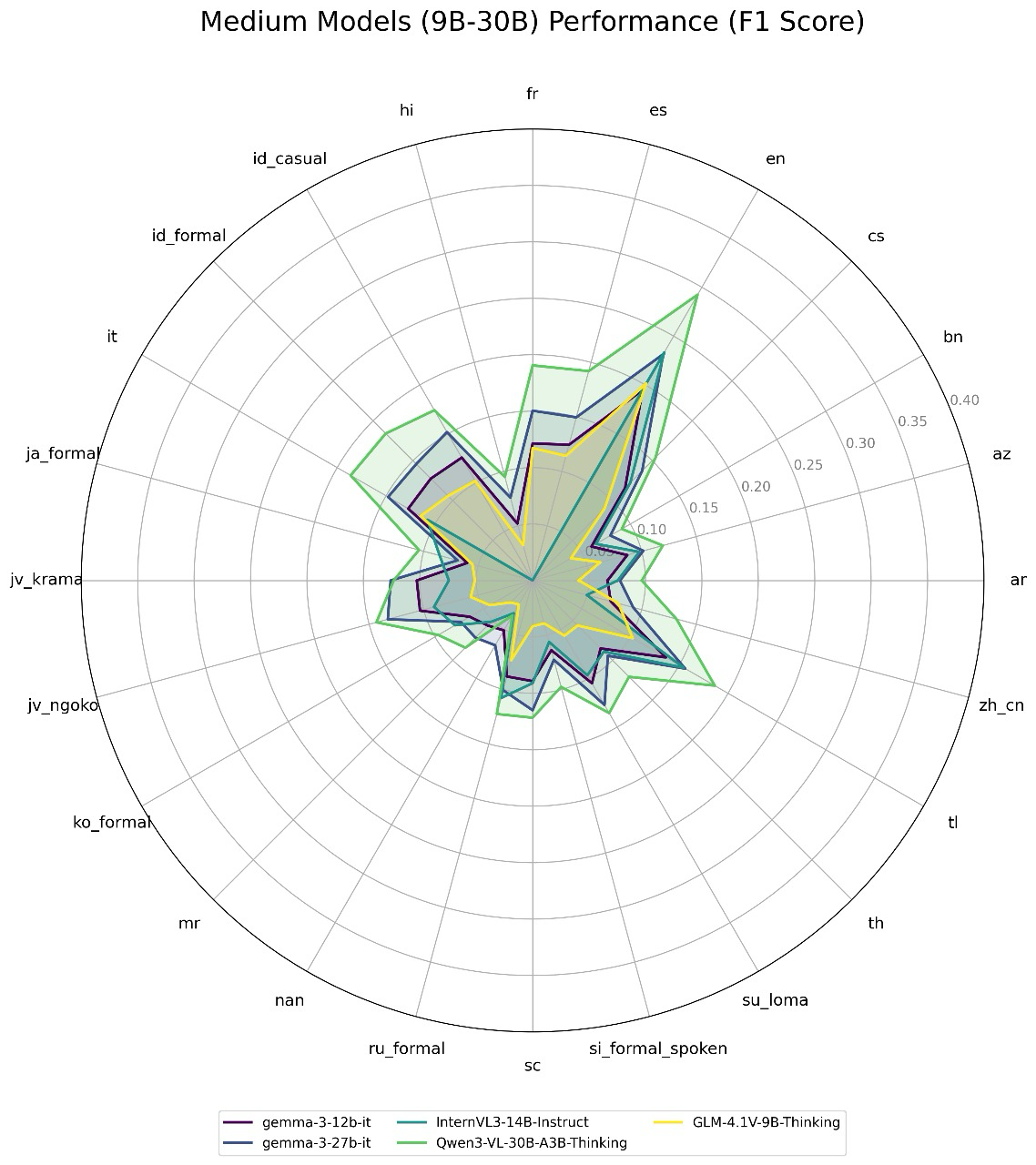}
    \caption{Medium (9B-30B)}
    \label{fig:radar_medium}
  \end{subfigure}
  \hfill
  \begin{subfigure}[b]{0.35\textwidth}
    \centering
    \includegraphics[width=\textwidth]{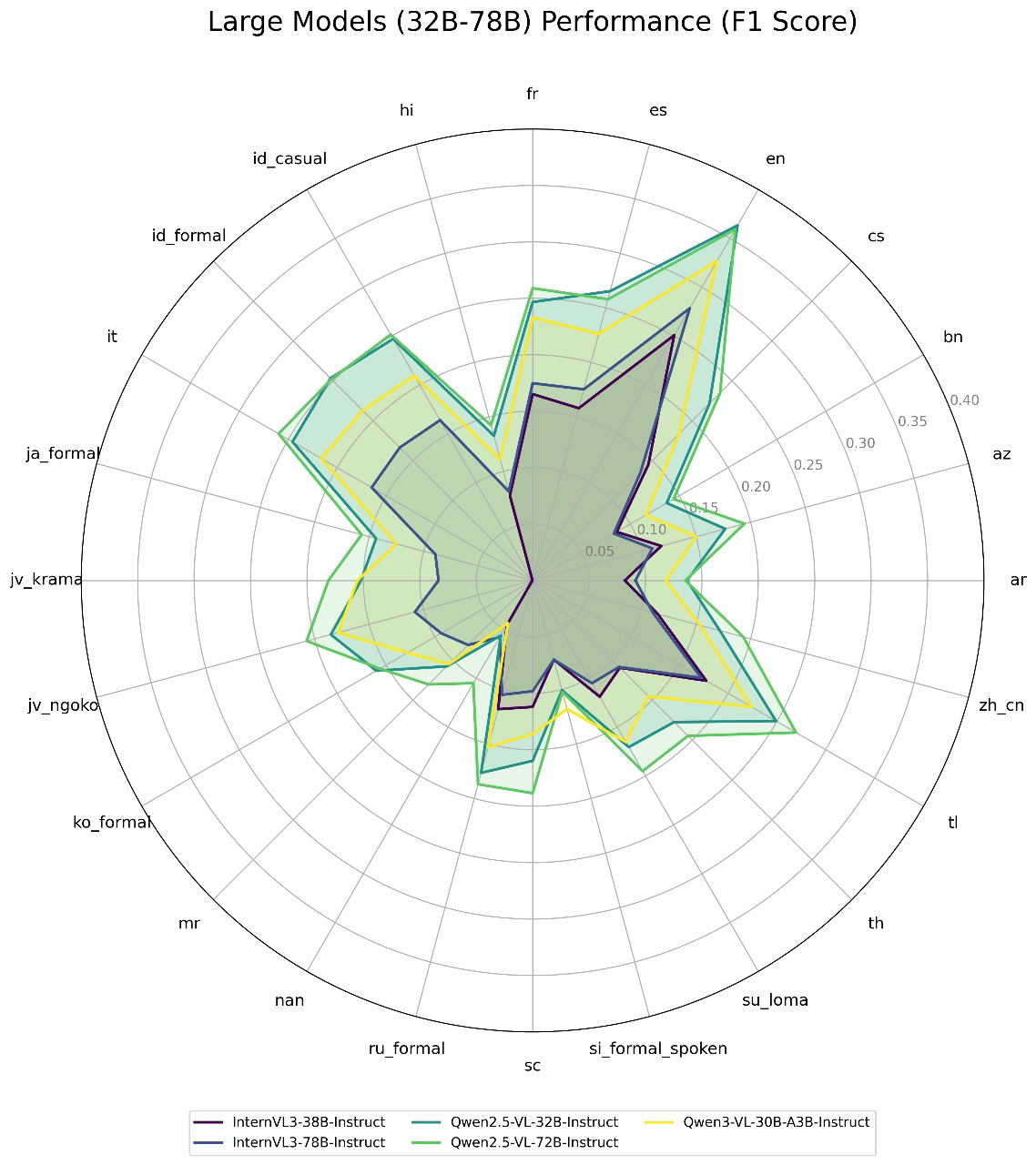}
    \caption{Large (32B-78B)}
    \label{fig:radar_large}
  \end{subfigure}
  \caption{Per-language F1 score performance on \textbf{MirageTVQA} across different model scales.}
  \label{fig:radar_plots}
\end{figure*}

\subsection{Visually-Rich and Realistic Rendering}
To address the reality gap, we developed a two-stage rendering pipeline. First, each table (in each language) is rendered into a clean PNG image from an HTML representation. We design 40+ distinct and advanced CSS themes to simulate a variety of document aesthetics, from minimalist and academic styles to financial reports and dark-mode interfaces. Next, we produce multiple noisy versions of each clean image. Using the \texttt{imgaug} library (\cite{imgaug}), we apply a stochastic pipeline of augmentations that are meant to simulate real-world degradation of documents. These involve geometric distortions, such as minor rotations, skew, and perspective transforms to simulate a camera capture; quality degradation, like Gaussian blur and variable JPEG compression; and the possibility of scanning artifacts, such as salt-and-pepper noise, slight scan lines, and corner shadowing. This process results in a rich dataset where each table is associated with one clean image and multiple unique noisy counterparts, complete with metadata for each applied transformation.

\subsection{Question-Answer Generation}
We produced question-answer pairs using a combined human-LLM approach to balance cost efficiency and quality of the annotation process. First, human annotators curated a seed corpus by manually creating one high-quality QA pair per table. Next, we expanded the QA dataset by utilizing the seed examples in detailed prompts, allowing LLMs (Google Gemini \citep{gemini_2024}) to produce 10 additional QA pairs for each table, covering 10 reasoning types and two question types. This process yielded 11 QA pairs for each table-language combination, for a total of 80,520 QA pairs (244 tables $\times$ 30 languages $\times$ 11 QA pairs). Following the generation of the QA pairs, we validated these pairs with LLMs and identified which pairs had been misclassified during generation. Three human annotators then corrected these flagged pairs. With this final correction, the QA pairs for each of the tables were complete. Both source tables and their associated QA pairs were translated into all target languages to enable extensive multilingual evaluation. For more details, refer to Appendix A.

\section{Results and Analysis}
Our benchmark, MirageTVQA, was designed to evaluate Vision-Language Models (VLMs) on two critical axes that reflect real-world challenges: \textbf{visual robustness} and \textbf{multilingual reasoning}. To do this, our dataset provides two visual settings for each table: digitally perfect clean images and noisy images that simulate document degradation. The analysis of model performance on MirageTVQA reveals significant and systematic limitations in current state-of-the-art models.

\subsection{Baseline Performance and Model Scale}
We first establish a performance baseline by evaluating models on the clean image set. The results, detailed in Table \ref{tab:language_results_full}, demonstrate a strong and consistent correlation between model scale and reasoning capability. The largest model evaluated, Qwen2.5-VL-72B (\cite{qwen2.5vl}), achieves the highest average Exact Match (EM) of 13.57\% across all languages, significantly outperforming smaller models. This trend, visually represented by the expanding area of the performance polygons in Figure \ref{fig:radar_plots}, confirms that greater parameter counts are beneficial for the complex, multi-step reasoning required by MirageTVQA.

\subsection{The Impact of Visual Noise}
A primary motivation for creating MirageTVQA was to measure how models performs with the visual imperfections common in real-world documents. Our findings reveal that current models are extremely brittle in this regard. Table \ref{tab:clean_vs_noise} directly compares model performance on `clean` versus `noisy` images for the English subset. The top-performing Qwen2.5-VL-72B (\cite{qwen2.5vl}) model, which scores 25.52\% EM on clean images, sees its performance degrade to just 16.50\% EM on the noisy set with drop of over \textbf{35\%}. This trend is consistent across all evaluated models. Hence, it validates a core premise of our benchmark: that performance on pristine, synthetic data is an unreliable predictor of performance on realistic, visually imperfect data.

\subsection{Multilingual Performance}
The second core motivation of MirageTVQA was to assess reasoning capabilities beyond English. Our results show the VLMs’ performance is biased towards English (see Figure \ref{fig:radar_plots}). Across all model scales, performance invariably peaks on English. Scores drop sharply even for other high-resource languages, degrade further for languages with different scripts, and become negligible for many low-resource languages. This demonstrates that current VLMs, despite their exposure to multilingual data during pre-training, fail to generalize complex, visually grounded reasoning skills to non-English contexts. This failure in cross-lingual transfer is a critical gap that MirageTVQA successfully exposes, reinforcing the need for more equitable and truly multilingual model development.

\begin{table*}[t!]
 \caption{Comprehensive per-language performance breakdown on \textbf{MirageTVQA}, showing Exact Match (\%). The `Avg.' column shows the overall average EM. I means Instruct models and T means Thinking models. Best result in each column is in \textbf{bold}.}
  \label{tab:language_results_full}
  \centering
  \resizebox{\textwidth}{!}{
  \setlength{\tabcolsep}{2pt}
  \renewcommand{\arraystretch}{0.7}
  \newcommand{\mysmallheader}[1]{\rotatebox{60}{\tiny #1}}
  
  \begin{tabular}{@{}llc|c|*{24}{c}@{}}
    \toprule
    \textbf{Size} & \textbf{Model} & \textbf{T} & \mysmallheader{Avg} & \mysmallheader{en} & \mysmallheader{es} & \mysmallheader{fr} & \mysmallheader{it} & \mysmallheader{cs} & \mysmallheader{ru} & \mysmallheader{tl} & \mysmallheader{id\_f} & \mysmallheader{id\_c} & \mysmallheader{su} & \mysmallheader{jv\_k} & \mysmallheader{jv\_n} & \mysmallheader{az} & \mysmallheader{th} & \mysmallheader{bn} & \mysmallheader{si} & \mysmallheader{ja} & \mysmallheader{ar} & \mysmallheader{sc} & \mysmallheader{zh} & \mysmallheader{hi} & \mysmallheader{ko} & \mysmallheader{mr} & \mysmallheader{nan} \\
    \midrule
    
    \parbox[t]{1mm}{\multirow{8}{*}{\rotatebox[origin=c]{90}{\tiny <10B}}} 
    & Pangea-7B (\cite{yue2024pangea})   & I & 1.03 & 5.23 & 1.29 & 1.29 & 1.36 & 3.62 & 1.00 & 0.92 & 1.31 & 1.27 & 0.80 & 0.62 & 0.83 & 0.71 & 0.66 & 0.41 & 0.21 & 0.25 & 0.36 & 0.37 & 0.62 & 0.49 & 0.41 & 0.21 & 0.12 \\
    & Qwen2.5-3B (\cite{qwen2.5})        & I & 1.83 & 5.19 & 2.75 & 2.90 & 2.38 & 2.10 & 2.37 & 2.05 & 2.88 & 2.63 & 1.39 & 1.44 & 1.53 & 1.13 & 2.06 & 1.15 & 0.82 & 1.44 & 0.81 & 1.10 & 1.32 & 1.51 & 1.60 & 1.11 & 0.12 \\
    & Gemma-3-4B (\cite{team2025gemma})  & I & 1.78 & 3.67 & 2.34 & 2.45 & 2.34 & 1.77 & 1.91 & 2.05 & 2.51 & 2.14 & 1.47 & 1.52 & 1.53 & 1.46 & 1.60 & 1.60 & 1.52 & 1.52 & 1.45 & 1.43 & 1.44 & 1.47 & 1.31 & 1.39 & 0.80 \\
    & Qwen2.5-7B (\cite{qwen2.5})        & I & 4.57 & 9.85 & 6.93 & 6.43 & 5.59 & 5.15 & 5.77 & 4.81 & 6.04 & 6.82 & 3.29 & 2.30 & 2.98 & 4.26 & 5.14 & 3.69 & 1.60 & 4.28 & 4.71 & 2.86 & 5.14 & 3.79 & 4.79 & 2.34 & 0.88 \\
    & Qwen3-8B (\cite{yang2025qwen3})     & I & 8.01 & 17.53 & 10.93 & 10.00 & 10.72 & 9.68 & 7.63 & 10.33 & 10.44 & 10.06 & 7.37 & 6.58 & 7.48 & 6.52 & 8.19 & 5.86 & 4.64 & 7.74 & 6.20 & 6.55 & 8.19 & 6.40 & 5.98 & 5.00 & 1.73 \\
    \cmidrule{2-28}
    & GLM-4.1V-9B (\cite{hong2025glm})    & T & 2.73 & 8.17 & 4.63 & 4.11 & 4.85 & 3.01 & 2.95 & 4.39 & 4.23 & 4.15 & 1.85 & 1.73 & 2.07 & 2.17 & 1.89 & 1.43 & 1.44 & 1.94 & 1.61 & 1.72 & 2.59 & 1.39 & 1.80 & 0.78 & 0.44 \\
    & Qwen3-8B (\cite{yang2025qwen3})     & T & 6.36 & 15.55 & 8.97 & 8.50 & 8.92 & 7.66 & 6.31 & 7.65 & 8.05 & 7.89 & 4.80 & 4.61 & 5.58 & 5.52 & 6.58 & 4.88 & 3.41 & 5.48 & 4.83 & 5.11 & 7.37 & 4.69 & 5.04 & 3.61 & 1.17 \\
    
    \midrule
    
    \parbox[t]{1mm}{\multirow{6}{*}{\rotatebox[origin=c]{90}{\tiny Med}}}
    & Gemma-3-12B (\cite{team2025gemma})  & I & 5.31 & 10.66 & 6.97 & 6.43 & 6.86 & 6.67 & 5.73 & 6.86 & 6.94 & 6.57 & 5.35 & 5.10 & 4.88 & 4.97 & 4.86 & 4.02 & 3.78 & 4.24 & 4.35 & 4.58 & 4.20 & 3.18 & 4.01 & 3.57 & 2.41 \\
    & InternVL3-14B (\cite{zhu2025internvl3}) & I & 3.69 & 12.72 & 0.00 & 0.00 & 5.26 & 6.34 & 6.31 & 7.19 & 0.00 & 0.00 & 3.92 & 2.63 & 3.72 & 4.68 & 3.91 & 3.40 & 2.71 & 5.68 & 4.23 & 4.09 & 3.37 & 0.00 & 4.06 & 2.54 & 1.13 \\
    & Gemma-3-27B (\cite{team2025gemma})  & I & 6.71 & 13.79 & 9.06 & 8.59 & 8.83 & 8.11 & 6.51 & 8.70 & 8.34 & 8.67 & 6.62 & 6.66 & 6.66 & 5.77 & 5.10 & 5.33 & 4.07 & 4.74 & 5.16 & 6.63 & 5.72 & 5.46 & 4.59 & 4.59 & 3.06 \\
    & Qwen3-30B (\cite{yang2025qwen3})    & I & 9.45 & 20.05 & 12.27 & 12.86 & 12.16 & 10.71 & 9.71 & 11.54 & 11.59 & 11.50 & 8.51 & 7.77 & 9.38 & 8.40 & 8.64 & 7.21 & 6.86 & 9.27 & 7.21 & 6.55 & 10.66 & 7.01 & 8.19 & 6.32 & 2.05 \\
    \cmidrule{2-28}
    & Qwen3-30B (\cite{yang2025qwen3})    & T & 8.19 & 18.95 & 11.44 & 11.07 & 11.18 & 9.18 & 7.76 & 9.82 & 10.48 & 10.14 & 6.87 & 6.33 & 7.85 & 6.90 & 7.33 & 5.82 & 5.92 & 7.37 & 5.92 & 6.91 & 9.26 & 6.40 & 6.02 & 5.33 & 1.85 \\
    
    \midrule
    
    \parbox[t]{1mm}{\multirow{3}{*}{\rotatebox[origin=c]{90}{\tiny Large}}}
    & InternVL3-38B (\cite{zhu2025internvl3}) & I & 4.76 & 15.16 & 9.31 & 9.25 & 9.16 & 8.48 & 7.55 & 10.20 & 8.50 & 8.34 & 6.15 & - & - & 7.02 & 6.01 & 5.49 & 3.90 & - & 4.67 & 5.97 & 7.61 & 4.93 & - & - & - \\
    & InternVL3-78B (\cite{zhu2025internvl3}) & I & 11.22 & 27.84 & 17.53 & 17.47 & 16.47 & 13.50 & 10.48 & 17.25 & 16.65 & 16.40& 10.51 & 8.30 & 10.80 & 10.98 & 10.83 & 8.31 & 7.26 & 8.9 & 9.16 & 9.80 & 11.02 & 8.16 & 9.35 & 8.08 & 5.80 \\
    & Qwen2.5-72B (\cite{qwen2.5})            & I & \textbf{13.57} & \textbf{25.52} & \textbf{17.57} & \textbf{17.42} & \textbf{18.08} & \textbf{15.94} & \textbf{14.07} & \textbf{17.52} & \textbf{17.01} & \textbf{16.84} & \textbf{12.22} & \textbf{10.98} & \textbf{12.86} & \textbf{13.21} & \textbf{13.50} & \textbf{10.66} & \textbf{6.54} & \textbf{11.90} & \textbf{10.07} & \textbf{12.15} & \textbf{14.28} & \textbf{10.80} & \textbf{11.27} & \textbf{9.39} & \textbf{5.47} \\
    \bottomrule
  \end{tabular}
  }
\end{table*}

\vspace{5pt}
\begin{table}[h!]
  \caption{Impact of Visual Noise on Model Performance (EM \%) on the English Subset. Models are sorted by their performance on clean images.}
  \label{tab:clean_vs_noise}
  \centering
  \begin{tabular}{lccc}
    \toprule
    \textbf{Model} & \textbf{Clean EM (\%)} & \textbf{Noisy EM (\%)} & \textbf{Perf. Drop (\%)} \\
    \midrule
    Gemma-3 27B-IT (\cite{gemma3}) & 13.79 & 12.87 & -6.7\% \\
    Qwen 2.5 VL 8B(\cite{qwen2.5vl}) & 17.53 & 16.62 & -5.2\% \\
    Qwen 2.5 VL 32B(\cite{qwen2.5vl})& 23.15 & 20.36 & -12.1\% \\
    \rowcolor{gray!10} 
    \textbf{Qwen 2.5 VL 72B (\cite{qwen2.5vl})} & \textbf{25.52} & \textbf{16.50} & \textbf{-35.3\%} \\
    \bottomrule
  \end{tabular}
\end{table}

\section{Conclusion}
In this work, we introduced MirageTVQA, a new large-scale benchmark designed to address a critical gap in the evaluation of vision-language models: their ability to reason over tables that are both visually imperfect and multilingual. Through our extensive experiments, we have demonstrated two significant findings. First, the performance of even state-of-the-art VLMs degrades in the presence of realistic visual noise, highlighting a profound lack of robustness. Second, we identified a severe "English-first" bias, with models failing to transfer their reasoning capabilities to non-English and low-resource languages. We hope it will be useful for the development of more robust and capable models for real-world table understanding. Additionally, we discuss outline future directions toward improving the interpretability of VLM failure modes in the Appendix.

\section{Limitations}
While this research provides an understanding of how real-world noise affects the performance of VLMs in multi-modal visual question-answers, we would like to emphasize a few limitations. First, we did analyze how noise impacts performance, but we did not establish interpretability methods to explain why these degradations occur or suggest ways to reduce that degradation. This will be an important area for future work. Second, we carried out our cross-lingual experiments in a limited scope. The cross-lingual experiments involved performance in only 25 languages, we only evaluated open models, and we did not perform experiments on top-tier proprietary models that may behave differently on robustness in noise. Adopting state-of-the-art proprietary models, introducing a wider array of languages, and exploring interpretability methods for understanding and addressing degradation are all potential improvements to our analyses.

\bibliographystyle{plainnat}
\bibliography{references}

\newpage

\appendix

\section{Technical Appendices and Supplementary Material}

\subsection{Data distribution and Language Composition}

To provide a transparent overview of our benchmark, Table \ref{tab:appendix_unified_stats} details the composition of MirageTVQA. Part (a) shows the initial breakdown of QA pairs and unique tables collected from our diverse sources before filtering and translation. Part (b) shows the final distribution of the nearly 60,000 high-quality QA pairs that passed our validation pipeline, broken down by language and grouped by linguistic family. This highlights the broad and balanced coverage of our final benchmark.

\begin{table*}[htbp]
  \caption{Detailed composition of the \textbf{MirageTVQA} benchmark. The table shows the distribution of tables and QA pairs by their original source, followed by the final distribution of QA pairs per language, grouped by linguistic family.}
  \label{tab:appendix_unified_stats}
  \centering
  \small 
  
  \begin{tabular}{@{}lllr@{}}
    \toprule
    \multicolumn{4}{c}{\textbf{(a) Data Source Composition}} \\
    \midrule
    \textbf{Data Source} & \textbf{\# QA Pairs} & \textbf{Unique Tables} & \textbf{Total Table Images} \\
    \midrule
    ArXiv & 12,100 & 1,100 & 4,400 \\
    Wikipedia & 27,500 & 2,500 & 10,000 \\
    Other (Financial, etc.) & 27,500 & 2,500 & 10,000 \\
    \midrule
    \rowcolor{gray!10}
    \textbf{Total Initial Set} & \textbf{67,100} & \textbf{6,100} & \textbf{24,400} \\
    \midrule[\heavyrulewidth] 
    \multicolumn{4}{c}{\textbf{(b) Language Composition (Final Valid Set)}} \\
    \midrule
    \textbf{Language Family} & \textbf{Language} & \textbf{\# Final QA Pairs} & \\ 
    \midrule

    Afro-Asiatic & Arabic (ar) & 2,482 & \\
    \midrule
    
    \multirow{6}{*}{\parbox{2.5cm}{\centering Austronesian}} 
    & Indonesian (id\_casual) & 2,435 & \\
    & Indonesian (id\_formal) & 2,434 & \\
    & Javanese (jv\_krama) & 2,431 & \\
    & Javanese (jv\_ngoko) & 2,419 & \\
    & Sundanese (su\_loma) & 2,373 & \\
    & Tagalog (tl) & 2,392 & \\
    \midrule
    
    \multirow{11}{*}{\parbox{2.5cm}{\centering Indo-European}} 
    & Bengali (bn) & 2,440 & \\
    & Czech (cs) & 2,428 & \\
    & English (en) & 2,618 & \\
    & French (fr) & 2,411 & \\
    & Hindi (hi) & 2,454 & \\
    & Italian (it) & 2,434 & \\
    & Marathi (mr) & 2,438 & \\
    & Russian (ru\_formal) & 2,410 & \\
    & Sardinian (sc) & 2,444 & \\
    & Sinhala (si\_formal\_spoken) & 2,433 & \\
    & Spanish (es) & 2,396 & \\
    \midrule

    Japonic & Japanese (ja\_formal) & 2,428 & \\
    \midrule

    Koreanic & Korean (ko\_formal) & 2,441 & \\
    \midrule

   Kra-Dai & Thai (th) & 2,430 & \\
    \midrule

    \multirow{2}{*}{\parbox{2.5cm}{\centering Sino-Tibetan}}
    & Hokkien (nan) & 2,487 & \\
    & Chinese (zh\_cn) & 2,430 & \\
    \midrule
    
    Turkic & Azerbaijani (az) & 2,392 & \\
    \midrule
    
    \rowcolor{gray!10}
    \textbf{Total Valid Set} & \textbf{24 Languages} & \textbf{58,480} & \\
    \bottomrule
  \end{tabular}

\end{table*}

\subsection{Table Translation Prompt}
For reproducibility, we provide the exact prompts used in our data generation pipeline. Figure \ref{fig:qa_generation_prompt} shows the detailed prompt provided to Gemini 1.5 Pro (\cite{comanici2025gemini}) to generate the English dense-reasoning question-answer pairs described in Section 3.3. Figure \ref{fig:translation_prompt} shows the prompt used to translate these QA pairs into the 25 target languages.

The following prompt shown in Fig \ref{fig:translation_prompt} was provided to the LLMs QA translation agents during the QA translation phase described in Section 3.1.


\begin{figure}[htbp]
\centering
\footnotesize
\setlength{\fboxsep}{3pt}
\setlength{\fboxrule}{0.3pt}
\fbox{%
\begin{minipage}{0.9\linewidth}
\ttfamily
You are an expert linguist and professional translator with deep expertise in structured data. Your task is to accurately translate a question-answer pair from English to \textbf{\{target\_language\}}.

The question-answer pair is based on the following data table. Use this table to understand the context of entities, numbers, and technical terms.

\textbf{Context Table:}

\{context\_table\_json\}

--------------------------------

\textbf{ENGLISH QUESTION-ANSWER PAIR TO TRANSLATE:}

\{english\_qa\_json\}

--------------------------------

\textbf{CRITICAL INSTRUCTIONS:}

1. Convert the question text into fluent and natural-sounding \textbf{\{target\_language\}}.

2. If the question\_type is \textbf{'open\_ended\_reasoning'}, you must translate the full text of the answer. However, if the question\_type is \textbf{'value'}, you must preserve the original answer exactly. Do not translate numbers (e.g., "370"), names (e.g., "Beta"), percentages, or codes.

3. Your entire response must be a single, valid JSON object with ONLY two keys: translated\_question and translated\_answer. Do not add any other text, explanations, or markdown code blocks.

------------------------------

\textbf{EXAMPLE:}

If target\_language is \textbf{Spanish} and the input QA pair is:

\{

\hspace*{1em}"question": "Which product experienced a decline in units sold from 2022 to 2023?",

\hspace*{1em}"answer": [["Beta"]],

\hspace*{1em}"question\_type": "value"

\}

Your perfect JSON output would be:

\{

\hspace*{1em}"translated\_question": "¿Qué producto experimentó una disminución en las unidades vendidas de 2022 a 2023?",

\hspace*{1em}"translated\_answer": [["Beta"]]

\}

Notice how "Beta" was NOT translated because the question\_type was 'value'.

---

\textbf{YOUR TASK:}

Translate the provided English QA pair to \textbf{\{target\_language\}} following all instructions.

\textbf{Your JSON Output:}
\end{minipage}%
}
\caption{LLM prompt for multilingual QA pair translation. Placeholders like \texttt{\{target\_language\}} and \texttt{\{context\_table\_json\}} represent actual input data provided to the model.}
\label{fig:translation_prompt}
\end{figure}

\begin{figure}[htbp]
\centering
\small
\setlength{\fboxsep}{6pt}
\setlength{\fboxrule}{0.5pt}
\fbox{%
\begin{minipage}{0.92\linewidth}
\ttfamily
You are a world-class data analyst and expert curriculum designer.

Your task is to generate a set of \textbf{\{num\_questions\}} diverse, challenging, and high-quality question-answer pairs based on the provided data table in JSON format.

The questions must require deep reasoning and not be simple lookups.

\textbf{CRITICAL INSTRUCTIONS:}

1. Create questions that cover a wide range of the reasoning categories defined below. Do not repeat question patterns.

2. Every question must be answerable \textit{exclusively} from the provided table. Do not require external knowledge.

3. \textbf{Provide Precise Answers:}

\hspace*{1em}- For \textbf{value-based questions}: The 'answer' must be the exact value(s) from the table or calculated from it. Format it as a list of lists (e.g., [["150"]] or [["Alpha"], ["Beta"]]).

\hspace*{1em}- For \textbf{open-ended reasoning questions}: The 'answer' should be a comprehensive explanation or analysis based on the data, formatted as a list containing a single list with one string element (e.g., [["The data shows a declining trend because..."]]).

4. Each question must have a 'question\_type' field that is either "value" or "open\_ended\_reasoning".

5. The 'evidence\_cells' must accurately list all cells needed to formulate the answer. Use standard spreadsheet notation (e.g., A1 for the top-left-most cell in the data body, where Column A is the first column and Row 1 is the first data row. Headers are considered Row 0, so A1 refers to the first data cell, not a header).

6. Your response MUST be a single, valid JSON object that conforms to the provided schema. Do not include any explanatory text, markdown, or comments outside of the JSON structure, and do not wrap the JSON in markdown code blocks (e.g., ```json).

\textbf{REASONING CATEGORIES TO USE:} Comparative Reasoning, Numerical Aggregation, Multi-Hop Reasoning, Temporal Reasoning, Conditional Reasoning, Proportional/Ratio Analysis, Hypothetical Reasoning, Correlation Inference, Structural/Metadata Reasoning, Outlier Detection

\textbf{REQUIRED JSON OUTPUT SCHEMA:}

\{

\hspace*{1em}"qa\_pairs": [

\hspace*{2em}\{

\hspace*{3em}"question": "string",

\hspace*{3em}"answer": [["string"]],

\hspace*{3em}"evidence\_cells": ["string"],

\hspace*{3em}"reasoning\_category": "string (must be one of the 10 categories)",

\hspace*{3em}"question\_type": "string (either 'value' or 'open\_ended\_reasoning')"

\hspace*{2em}\}

\hspace*{1em}]

\}

NOW, GENERATE QA PAIRS FOR THE FOLLOWING TABLE:

Input Table (as JSON): \{table\_as\_json\_string\}

Your JSON Output:
\end{minipage}%
}
\caption{LLM prompt for automated QA pair generation. Placeholders like \texttt{\{table\_as\_json\_string\}} represent the actual table data provided to the model.}
\label{fig:qa_generation_prompt}
\end{figure}

\end{document}